Preprint

# Using a Large Language Model to generate a Design Structure Matrix

Edwin C.Y. Koh

Design and Artificial Intelligence Programme & Engineering Product Development Pillar
Singapore University of Technology and Design, Singapore, Singapore

**Abstract**

The Design Structure Matrix (DSM) is an established method used in dependency modelling, especially in the design of complex engineering systems. The generation of DSM is traditionally carried out through manual means and can involve interviewing experts to elicit critical system elements and the relationships between them. Such manual approaches can be time-consuming and costly. This paper presents a workflow that uses a Large Language Model (LLM) to support the generation of DSM and improve productivity. A prototype of the workflow was developed in this work and applied on a diesel engine DSM published previously. It was found that the prototype could reproduce 357 out of 462 DSM entries published (i.e. 77.3%), suggesting that the work can aid DSM generation. A no-code version of the prototype is made available online to support future research.

**Contact:** edwin_koh@sutd.edu.sg

## 1. Introduction

The Design Structure Matrix (DSM) is an analysis tool for system modelling [Steward 1981] and has been applied in various areas including the modelling of product, organisation, and process architecture [Eppinger and Browning 2012]. DSM is known for its simplicity and conciseness in representation and exists in the form of a $N \times N$ square matrix that maps the relationships between the set of $N$ system elements [Yassine and Braha 2003; Browning 2015]. An example DSM ($N = 4$) is shown in Figure 1. Based on the DSM convention described by Browning [2001], Element 1 depends on Element 2 as indicated by a red cell entry in row 2 column 1 of the DSM. Likewise, Element 4 depends on Element 3 as indicated in row 3 column 4. The diagonal of the DSM maps each element to itself and is indicated as black cells in Figure 1. The diagonal is usually left empty but is sometimes used as a space to store element-specific data, such as the likelihood of changing the given element based on market projection [Koh et al. 2013]. The DSM in Figure 1 is not symmetrical across the diagonal, indicating asymmetrical dependencies between the system elements. For example, Element 1 depends on Element 2 but Element 2 does not depend on Element 1. In contrast, the example DSM shows that Element 2 and Element 4 have a symmetrical interdependency. It is important to note that a transposed version of the DSM convention is also widely adopted by many (e.g. Clarkson et al. [2004]; Koh et al. [2015]). Nevertheless, regardless of the convention used, the set of $N$ system elements and the relationships between them are necessary data to be collected for DSM generation.

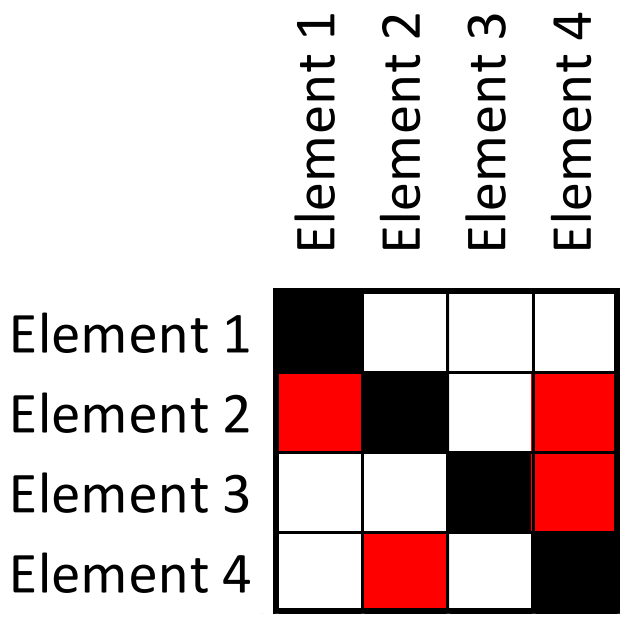

**Figure 1: An example DSM.**

A series of DSM use cases documented in [Eppinger and Browning 2012] reveals that data collection is traditionally carried out through surveys and interviews. For instance, Clarkson, Simons, and Eckert [2012, pp58] present a DSM use case at an industry partner and stated that chief engineers, deputy chief engineers, and 17 senior engineers were involved in providing inputs to generate a DSM of a helicopter for design change evaluation. Suh and de Weck [2012, pp 43] describe another DSM use case at a different organisation and indicated that 140 person hours were spent in creating a DSM of a digital printing system consisting of 84 components. Sosa et al. [2012, pp 105] discuss the use of an organisation architecture DSM at an aerospace company to identify communication patterns between and within design teams and revealed that a four-month period was used to interview lead engineers of different teams to generate the DSM. The examples underline how time-consuming and costly it can be to put together a DSM manually.

This paper presents a workflow that uses a Large Language Model (LLM) to support the generation of DSM. The proposed workflow, referred to as Auto-DSM in this paper, identifies DSM headings and populates DSM entries by querying organisation-specific proprietary data with an LLM. The goal is to improve productivity through automation. A prototype was developed and applied in this work to examine the feasibility of the workflow. The results were compared with those derived directly from a python implementation of ChatGPT (https://platform.openai.com/docs/models/gpt-3-5) to examine if the proposed workflow add value. For simplicity, further mention of ChatGPT in this work refers to the python implementation. A no-code version of the workflow was developed and made available online to support independent testing and future research. The main contributions of this paper are summarised as follows:

(1) A workflow for automated DSM generation
(2) Analysis results on the feasibility of the proposed workflow
(3) A no-code prototype version of the proposed workflow (available at: https://drive.google.com/file/d/18TIXWhzZGz6z9YSgVEFIAnB1qSUcpAH0/view?usp=sharing)

## 2. Previous work

Applications of Design Structure Matrix (DSM) are well reported in research publications. For example, Eppinger et al. [1994] describe how the sequencing of design tasks can be carried out through a DSM to reduce iterations and shorten development time. Clarkson et al. [2004] present the use of a DSM approach to analyse and predict change propagation between product components. Maier et al. [2008] describe a method to analyse the perception of communication between and within teams using a DSM. Lin et al. [2012] explore how manufacturing activities can be scheduled through DSM to reduce uncertainty. De Lessio et al. [2019] discuss the planning system and how stakeholders and documents generated during design and development can be analysed and aligned through a DSM. Koh [2022a] describes a DSM approach to map asymmetrical indirect dependencies between infrastructure system elements for resilience planning. Moon and Suh [2023] introduce a method that uses a DSM to assess the economic feasibility of infusing multiple technologies into the next product system. Indeed, the DSM has been used in a wide range of modelling tasks as acknowledged in a dedicated and broad review of DSM applications put together by Eppinger and Browning [2012]. Further readings on DSM are also available in topic-specific reviews, such as the review on process improvement by Wynn and Clarkson [2018] and the review on design change management by Brahma and Wynn [2023]. A detailed review on DSM is therefore not reproduced here. Instead, this section focuses the discussion on the automation of DSM generation.

The pioneering work of Dong and Whitney [2001] is recognised as the first to attempt the automation of DSM generation (see [Wilschut et al. 2018]). The work builds on axiomatic design theory [Suh 1998] to generate DSMs by extracting the relationships between design parameters and functional requirements. While the method demonstrates the feasibility of automating DSM generation based on relationships derived through graphical representations, Tosserams et al. [2010] suggest that language-based approaches can provide better scalability. Indeed, the benefits of using natural language processing to support design analysis is widely documented [Chiu et al. 2023].

A method that specifically uses a language-based approach for DSM generation was introduced by Wilschut et al. [2018], which utilises design-related text written in a specific grammatical structure by trained personnel. The work highlights the potential of using natural language in DSM generation and offers insights on how similar research can be adapted. For example, Kang and Tucker [2016] present an approach to quantify dependencies between system elements by mining textual technical descriptions from a textbook. Akay and Kim [2020] demonstrate how functional dependencies between design parameters can be extracted through deep learning language models. Sarica et al. [2020] present a technology semantic network (TechNet) that builds on patent data to support the retrieval of unique relational knowledge between entities. These language-based methods were not applied in DSM generation but have the potential to be extended to establish the dependencies between system elements in a DSM. An extensive review on the use of natural language processing in design is available in [Siddharth et al. 2022]. To date, the utilisation of a Large Language Model (LLM) to generate a DSM has not yet been documented.

### 3. Auto-DSM: A workflow based on an LLM to automate DSM generation

Large Language Models (LLM) can be trained from scratch and finetuned to suit design research [Siddharth et al. 2022]. However, there are sustainability concerns such as the carbon footprint associated to the training and finetuning of LLMs [Wynsberghe 2021]. Hence, this work explores the use of an off-the-shelf LLM to query organisation-specific proprietary data for DSM generation. The utilisation of an off-the-shelf LLM also implies that the results generated can improve over time simply by plugging in a better version when available, making the approach even more sustainable. The entire workflow is illustrated in Figure 2.

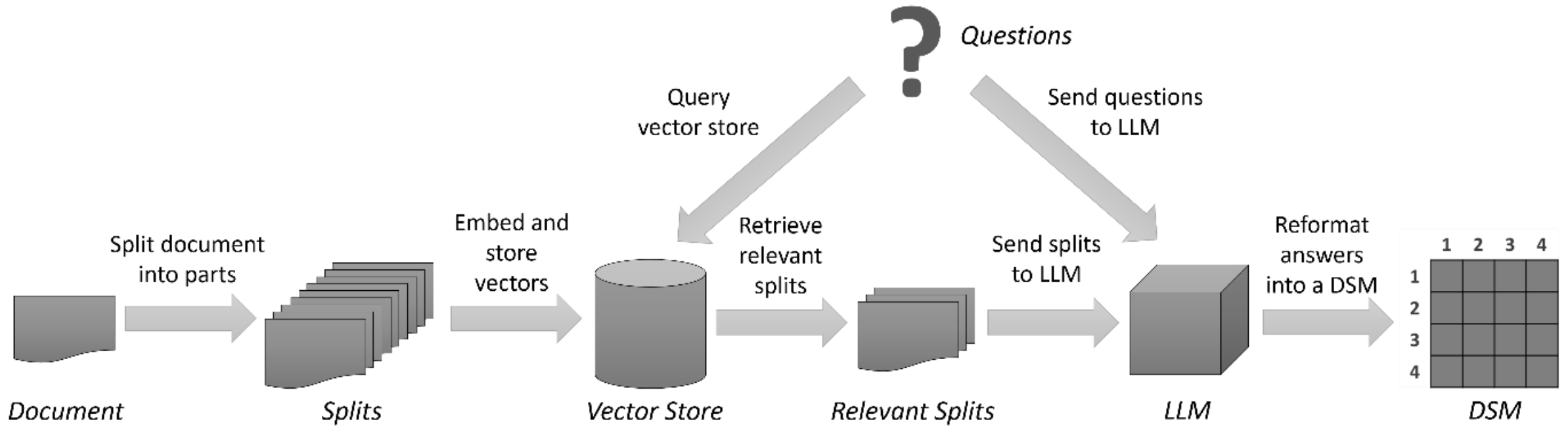

**Figure 2: A workflow that uses an LLM to establish DSM headings and entries based on organisation-specific proprietary data.**

In the first instance, relevant data in the form of a document is split into smaller parts for text embedding and stored in a vector store. A set of 2 predefined questions pertinent to DSM generation will subsequently be used to query the vector store to retrieve relevant splits that may contain answers to the questions posed. The first question queries the vector store to identify the system elements of the system analysed – "*Identify the main {product} components that make up a {product}. Output the answer as a list and a python list. Do not output anything else. If you don't know the answer, strictly state **I don't know** instead of making up an answer*." The variable *{product}* can be any product system of interest to be inserted into the question (e.g. replace with "engine" to create a DSM of an engine). Once the list of system elements is identified, each element will be sequentially fed into the second question to query the relationships between the system elements found – "*Are {Element A} and {Element B} {linkage-type} linked? State **Yes** or **No**. If you don't know the answer, strictly state **I don't know** instead of making up an answer.*" For this question, the entries for *{Element A}* and *{Element B}* are systematically inserted based on the list derived from the first question. The variable *{linkage-type}* can be defined to specifically examine a linkage type (e.g. replace with "mechanically" to examine mechanical links) or left empty to identify unspecified links. Finally, a DSM is constructed based on the answers derived from the first question (i.e. DSM headings) and the answers derived from the second question (i.e. DSM entries).

In this work, a no-code executable (.exe) prototype was developed to demonstrate and examine the proposed workflow. The LangChain library 'langchain.text_splitter' is used for text splitting, where each split is set as 1000 characters with an overlap of 150 characters with the next split (see https://python.langchain.com/docs/get_started/introduction.html for further information on LangChain). Next, embedding will be done using an OpenAI Large Language Model (LLM) through 'langchain.embeddings.openai' and stored in a vector store through 'langchain.vectorstores'. Splits that are relevant to the 2 predefined questions will be retrieved using the 'RetrievalQA' function from 'langchain.chains', where retrieval is based on 'similarity' and four splits will be retrieved in each retrieval (i.e. same as the default setting). The retrieved splits and the questions are processed using the 'ChatOpenAI' function from 'langchain.chat_models', which will access the OpenAI 'gpt-3.5-turbo' LLM with 'temperature' set as '0' to produce near identical results in repeated queries. The results produced (i.e. DSM headings and entries) will subsequently be arranged in a DSM format. The DSM headings identified will be kept unmodified. However, the DSM entries produced will be converted with '1' representing a ***Yes*** answer from the second question, '0' representing ***No***, and '5' representing ***I don't know***. All diagonal entries will be assigned as '1'. Lastly, a comma-separated values (.csv) file of the DSM generated will be created as shown in Figure 3.

| | Thermally | Heat pum | Compress | Condense | Evaporato | Expansion | Refrigeran | Temperat | Door(s) | Shelves or |
|---|---|---|---|---|---|---|---|---|---|---|
| Thermally | 1 | 0 | 0 | 0 | 0 | 0 | 0 | 0 | 0 | 0 |
| Heat pum | 0 | 1 | 0 | 0 | 0 | 0 | 1 | 0 | 0 | 0 |
| Compress | 0 | 1 | 1 | 1 | 0 | 0 | 1 | 0 | 0 | 0 |
| Condense | 0 | 0 | 1 | 1 | 0 | 0 | 0 | 0 | 0 | 0 |
| Evaporato | 0 | 0 | 0 | 0 | 1 | 0 | 0 | 0 | 0 | 0 |
| Expansion | 0 | 0 | 0 | 0 | 0 | 1 | 0 | 0 | 0 | 0 |
| Refrigeran | 0 | 5 | 1 | 0 | 0 | 1 | 1 | 0 | 0 | 0 |
| Temperat | 0 | 0 | 0 | 0 | 0 | 0 | 0 | 1 | 0 | 0 |
| Door(s) | 0 | 0 | 0 | 0 | 0 | 0 | 0 | 0 | 1 | 0 |
| Shelves or | 0 | 0 | 0 | 0 | 0 | 0 | 0 | 0 | 0 | 1 |

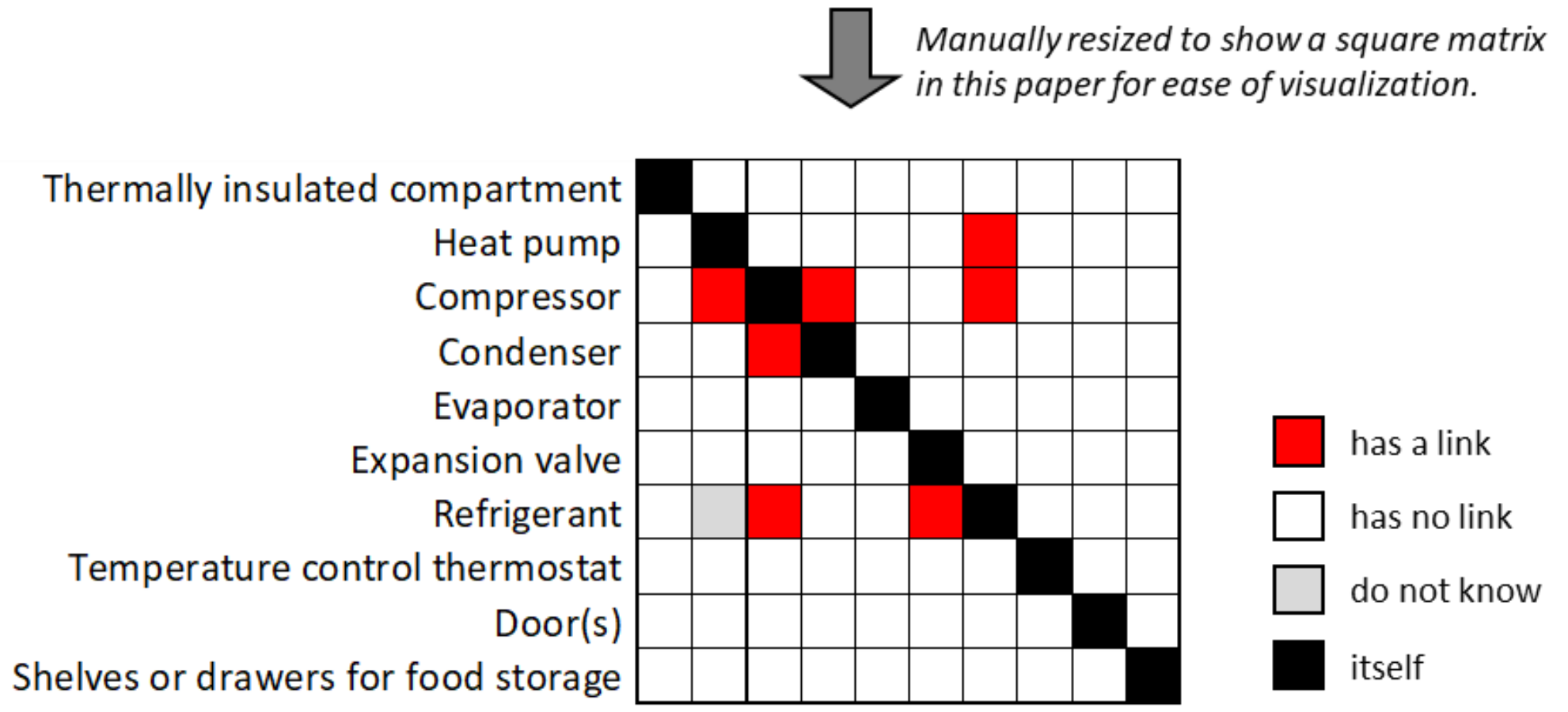

**Figure 3: An example output of the prototype – original csv version (top) and resized square version in this paper (bottom).**

## 4. A test case: Applying Auto-DSM on a diesel engine example

This section documents a test case that examines the feasibility of using Auto-DSM in generating a DSM of a diesel engine. A diesel engine was used in this work as its components (i.e. system elements) and their interdependencies are well documented, allowing comparisons to be made. Five settings with different combinations of input data used in Auto-DSM were examined in this work. Setting 1 uses a handbook on the design and development of heavy-duty diesel engines [Lakshminarayanan and Agarwal 2020] as the input data for Auto-DSM to examine the scenario of using a rich input dataset. Setting 2 uses another handbook on diesel engines [Mollenhauer and Tschoeke 2010] as a comparison. Setting 3 uses 1,634 YouTube comments on diesel engines documented by [Koh 2022b] to examine the use of unstructured lower quality input data. Setting 4 examines the merits of expanding the size of input data by combining the two handbooks as the input data used [Lakshminarayanan and Agarwal 2020; Mollenhauer and Tschoeke 2010], while Setting 5 further appends YouTube comments documented by [Koh 2022b] to examine the effect of adding lower quality input data to the dataset. A python implementation of ChatGPT (gpt-3.5-turbo, temperature=0) was also used in this work to establish if there was value-add in using Auto-DSM instead of using an LLM directly. A manually generated diesel engine DSM published by [Keller et al. 2005] was used in this work as well to serve as a reference for the test case. For ease of reading, the input data used in this paper are numbered as shown in Table 1.

**Table 1: Data used in this work.**

| | |
|---|---|
| [1] | Lakshminarayanan, P.A. and Agarwal, A.K. (2020) Design and Development of Heavy Duty Diesel Engines: A Handbook. |
| [2] | Mollenhauer, K. and Tschöke, H. (2010) Handbook of Diesel Engines. |
| [3] | Koh, E.C.Y. (2022) Design change prediction based on social media sentiment analysis, AI EDAM Vol. 36. |
| [4] | Keller, R., Eger, T., Eckert, C.M., and Clarkson, P.J. (2005) Visualising Change Propagation, ICED'05. |

### 4.1 Comparing DSM headings

As discussed previously, the components identified by Auto-DSM are used as row and column headings for the DSM generated. Table 2 shows the results on the engine components found by Auto-DSM based on the input data used. It can be seen that the number of components found can vary by more than three times from as few as 4 components (based on [3]) to as many as 15 components (based on [1, 2]). As mentioned, data in [3] are YouTube comments and are of lower quality compared to data in [1, 2] captured from two diesel engine handbooks. This suggests that the results generated are sensitive to the quality of the input data used in Auto-DSM.

**Table 2: Diesel engine components found based on different input data used in Auto-DSM.**

| | Input data used in Auto-DSM | | | | |
|---|---|---|---|---|---|
| **S/N** | **[1]** | **[2]** | **[3]** | **[1, 2]** | **[1, 2, 3]** |
| 1 | Air-cleaner | Crankshaft | Piston | Cylinder block | Crankshaft |
| 2 | Compressor | Cylinder block | Crankshaft | Cylinder head | Cylinder block |
| 3 | Turbocharger | Pistons | Valve | Pistons | Pistons |
| 4 | Intake manifold | Connecting rods | Oil lubricates | Connecting rods | Connecting rods |
| 5 | Air cooler | Cylinder head | | Crankshaft | Cylinder head |
| 6 | Fuel injectors | Valves | | Camshaft | Valves |
| 7 | Cylinders | Camshaft | | Valves | Camshaft |
| 8 | Exhaust gas turbine | Intake manifold | | Intake manifold | Intake manifold |
| 9 | ECU (Engine Control Unit) | Exhaust manifold | | Exhaust manifold | Exhaust manifold |
| 10 | Engine interface | Fuel injectors | | Fuel injectors | Fuel injectors |
| 11 | Monitor | Ignition system | | Ignition system | Ignition system |
| 12 | | Cooling system | | Cooling system | Cooling system |
| 13 | | Lubrication system | | Lubrication system | Lubrication system |
| 14 | | | | Timing belt/chain | |
| 15 | | | | Flywheel | |

The number of components found fell to 13 when [3] was added to [1, 2] to form a dataset of [1, 2, 3] for Auto-DSM. Interestingly, the 13 components found and the order that they appeared are identical to the results generated using [2] (i.e. a high-quality dataset). This suggests that the presence of low-quality data in a high-quality dataset has a smaller influence on the results (i.e. contrast [1, 2] with [1, 2, 3]) compared to the exclusive use of low-quality data (i.e. contrast [1, 2] with [3]).

The same prompts used in Auto-DSM were subsequently used on ChatGPT to examine how the results might differ if ChatGPT was used directly. 20 engine components were found (see Table 3), which is more than the number of components found in each of the Auto-DSM settings examined (see Table 2). Although it appears that the direct use of ChatGPT can give more comprehensive results as more components were identified, it should be highlighted that the components identified through ChatGPT were not derived through organisation-specific proprietary data. For instance, with reference to [4] in Table 3, 22 engine components were identified by company experts in the context of their organisation. Examples of the components identified include 'Cylinder Head Assembly' and 'Flywheel & Ring Gear'. The naming of these components shows that the component boundaries at the organisation differs from the component boundaries used by ChatGPT (i.e. 'Cylinder head' and 'Flywheel'). This suggests that the ability to use proprietary data with organisation-specific details in Auto-DSM can produce results that are more relevant than using ChatGPT directly.

**Table 3: Diesel engine components identified using ChatGPT and through Experts.**

| S/N | ChatGPT (gpt-3.5-turbo, temperature = 0) | Expert inputs [4] |
|---|---|---|
| 1 | Cylinder block | Cylinder Head Assembly |
| 2 | Cylinder head | Cylinder Block Assembly |
| 3 | Pistons | Piston & Rings & Gudgeon Pin |
| 4 | Connecting rods | Connecting Rod |
| 5 | Crankshaft | Crankshaft & Main Bearings |
| 6 | Camshaft | Valve Train |
| 7 | Valves | Cam Shaft |
| 8 | Intake manifold | Push Rods |
| 9 | Exhaust manifold | High Pressure Fuel Pipes |
| 10 | Fuel injectors | Engine Control Module |
| 11 | Spark plugs | Fuel Pump |
| 12 | Timing belt/chain | Fuel Injection Assembly |
| 13 | Oil pump | Adapter Plate & Flywheel Housing |
| 14 | Water pump | Flywheel & Ring Gear |
| 15 | Alternator | Starter Motor |
| 16 | Starter motor | Sump |
| 17 | Flywheel | Oil Pump |
| 18 | Engine control unit (ECU) | Gear Train |
| 19 | Cooling system components (radiator, thermostat, fan) | Turbocharger |
| 20 | Lubrication system components (oil pan, oil filter, oil cooler) | Aircharge Cooler |
| 21 | | Exhaust Manifold |
| 22 | | Wiring Harness |

### 4.2 Comparing DSM entries

By using the Auto-DSM, the components identified in the previous section were automatically assigned as DSM headings for DSM entries generation. The design dependency examined in this paper was limited to mechanical links (i.e. physical contact). This implies that the DSM entries generated should be symmetrical across the diagonal (i.e. having a mechanical link in Row 1 Column 2 means having a mechanical link in Row 2 Column 1). In addition, as the prompt structure of Auto-DSM allows the generation of ***I don't know*** as an answer to avoid hallucination, not all entries ended up with a useful label indicating either ‘has a link’ or ‘has no link’ (‘do not know’ entries are not useful). Figure 4 shows the result of the DSM generated based on [1], with 11 components (DSM headings) and 110 DSM entries. Entries labelled as ‘has a link’ are coloured in red, entries labelled as ‘has no link’ are coloured in white, and entries labelled as ‘do not know’ are coloured in grey for ease of visualisation. The entire process (i.e. the workflow in Figure 2) took less than 4 minutes to complete using the no-code prototype on a 11th Gen Intel(R) Core(TM) i7-1185G7 3.00GHz processor with 16.0 GB installed RAM.

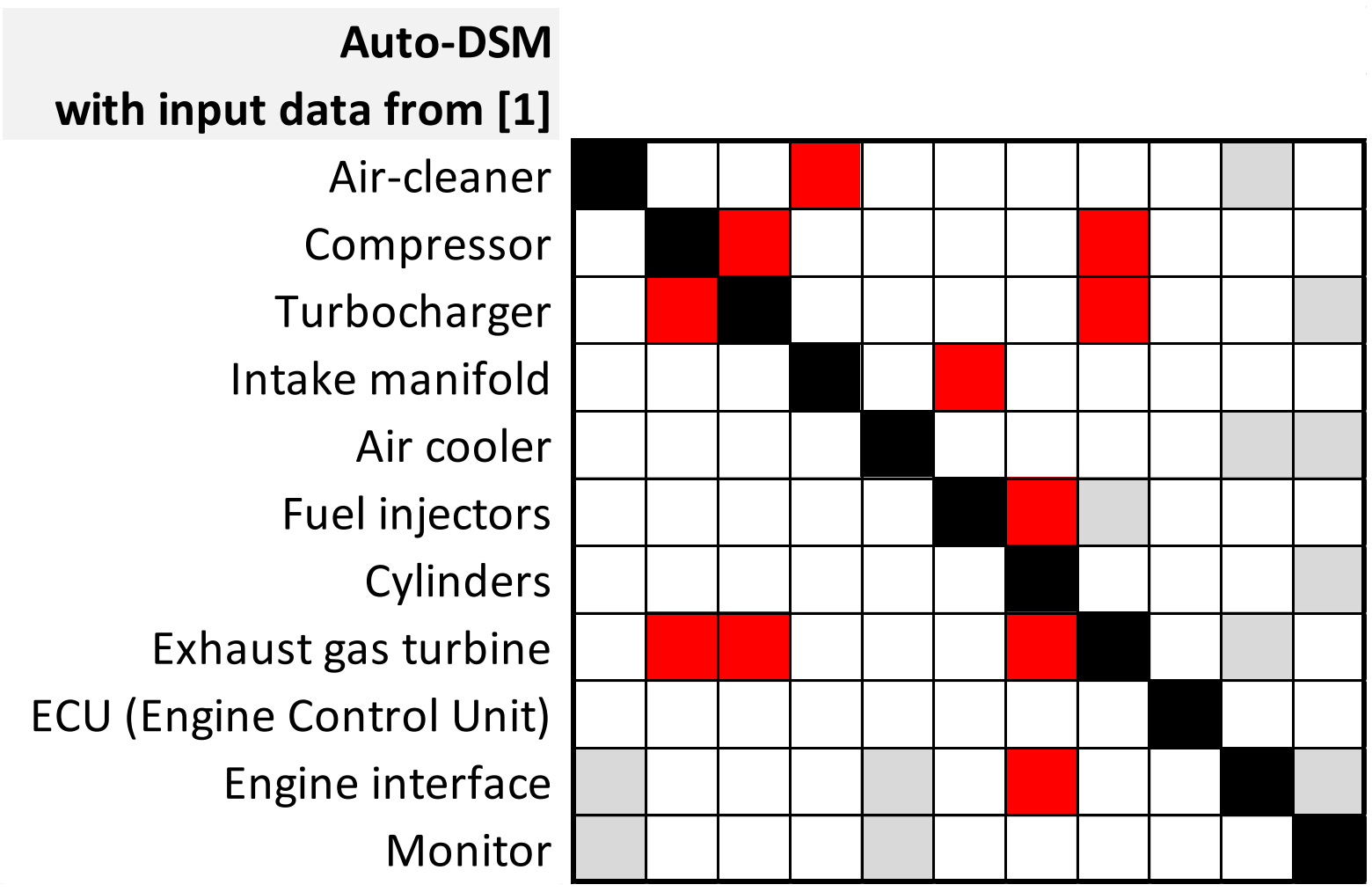

**Figure 4: DSM generated based on Auto-DSM with input data from [1].**

Two metrics were used in this work to evaluate the results based on the requirement to produce symmetrical entries and the possibility of generating a ‘do not know’ label. The first metric is *Correctness*, which is calculated as the number of symmetrical links found divided by the total number of links found, expressed as a percentage. The second metric is *Completeness*, which is calculated as the number of entries with a useful label (i.e. having a ‘has a link’ or a ‘has no link’ label) divided by the total number of DSM entries to be filled, expressed as a percentage. A summary of the results produced using different input data is provided in Table 4.

It can be seen from Table 4 that the percentage of symmetrical links (i.e. Correctness) can decrease by half depending on the input data used in Auto-DSM. For instance, Auto-DSM with handbook [2] produced a Correctness of 82.6% while Auto-DSM with YouTube comments [3] produced a Correctness of 40.0%. However, it is important to point out that the use of a high-quality dataset does not guarantee Correctness as demonstrated by Auto-DSM with handbook [1], which achieved a Correctness of 54.5%. In addition, combining input data in this work resulted in a pooling effect, where the Correctness for [1, 2] and for [1, 2, 3] were found to be within the highest and lowest Correctness value achieved by their constituent input data. The results also revealed that the percentage of entries with a ‘has a link’ or a ‘has no link’ label (i.e. Completeness) is not correlated to the percentage of symmetrical links found (i.e. Correctness). For example, Auto-DSM with [3] produced the lowest Correctness of 40% and has a Completeness of 100%, whereas Auto-DSM with [2] has the highest Correctness of 82.6% and has a lower Completeness of 99.4%. This suggests that Correctness and Completeness are two independent metrics.

**Table 4: A summary of results on DSM entries derived through Auto-DSM.**

| Input data used in Auto-DSM | Number of | | | | |
|---|---|---|---|---|---|
| | Components found | DSM entries to be filled¹ | Links found by Auto-DSM² | Symmetrical links found (as % of ²) | Entries with a useful label (as % of ¹) |
| [1] | 11 | 110 | 11 | 6 (54.5%) | 98 (89.1%) |
| [2] | 13 | 156 | 46 | 38 (82.6%) | 155 (99.4%) |
| [3] | 4 | 12 | 5 | 2 (40.0%) | 12 (100.0%) |
| [1, 2] | 15 | 210 | 57 | 40 (70.2%) | 208 (99.0%) |
| [1, 2, 3] | 13 | 156 | 47 | 34 (72.3%) | 154 (98.7%) |

Figure 5 shows the DSM generated based on the same components derived from Auto-DSM with [1], but with DSM entries generated through ChatGPT directly. The objective is to examine whether ChatGPT can produce better Correctness and Completeness compared to the Auto-DSM results produced. Similar to Figure 4, entries labelled as 'has a link' are coloured in red, entries labelled as 'has no link' are coloured in white, and entries labelled as 'do not know' are coloured in grey. A summary of the ChatGPT results based on the components derived through Auto-DSM is provided in Table 5.

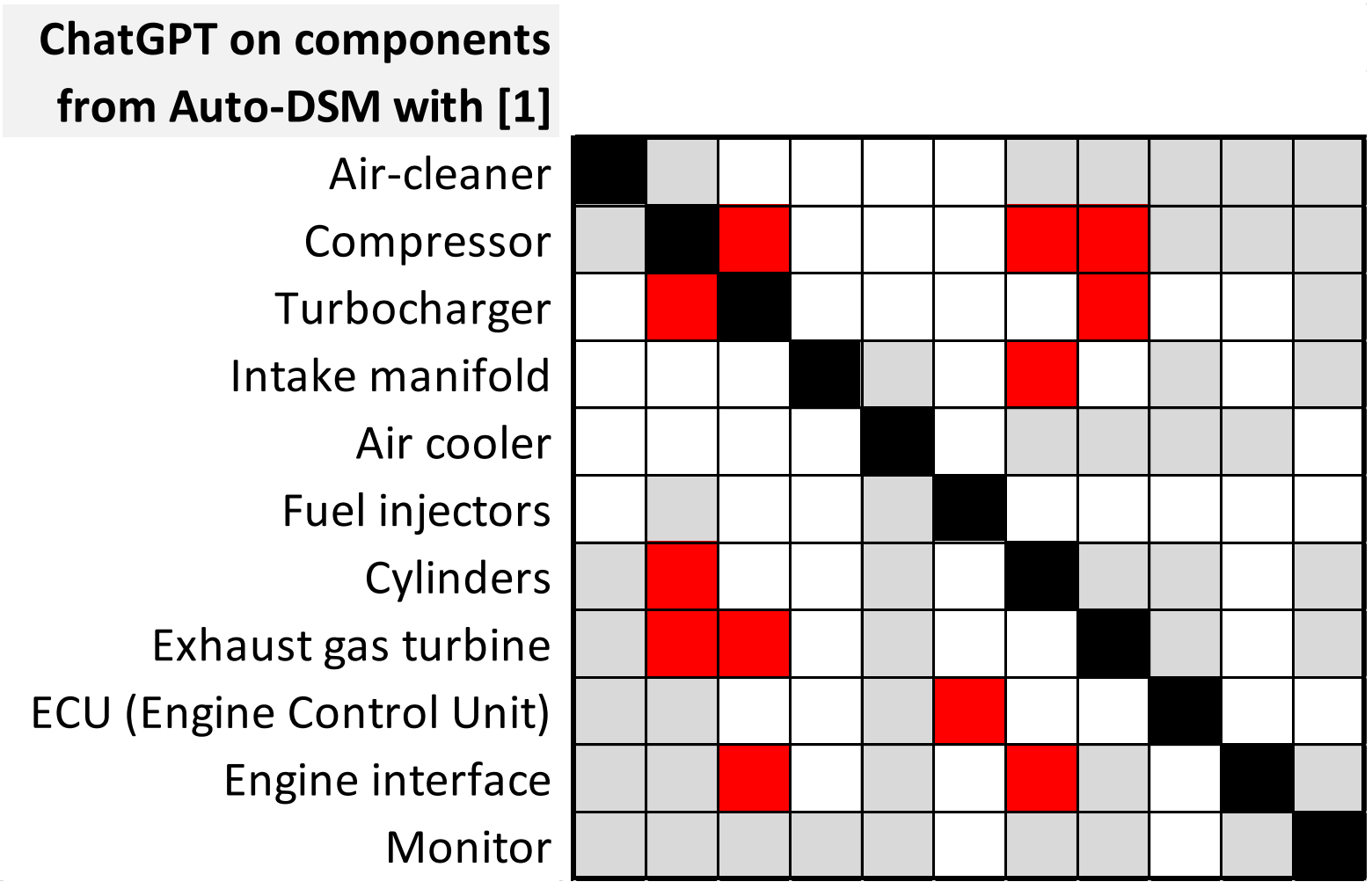

**Figure 5: Regeneration of Figure 4 with entries derived using ChatGPT.**

It can be seen from Table 5 that the percentage of symmetrical links (i.e. Correctness) varied between 66.7% and 85.7%, which is a smaller variation range compared to Table 4 where the range is between 40% and 82.6%. This suggests that the direct use of ChatGPT can produce results that are more consistent across DSM models. Indeed, the engine components identified in [2] and [1, 2, 3] are identical and resulted in the same Correctness of 70.0% when ChatGPT was used directly. ChatGPT produced higher Correctness for [1] and [3] but lower Correctness for [2], [1, 2], and [1, 2, 3] when compared to the results in Table 4. Hence, it was inconclusive if ChatGPT can produce better Correctness based on the results of this work. Nevertheless, Completeness was found to be lower across all the settings examined when ChatGPT was used directly, suggesting that the use of Auto-DSM in populating DSM entries has an advantage.

**Table 5: A summary of results on DSM entries derived through ChatGPT.**

| Input data used in Auto-DSM | Number of | | | | |
|---|---|---|---|---|---|
| | Components found | DSM entries to be filled[1] | Links found by ChatGPT[3] | Symmetrical links found (as % of [3]) | Entries with a useful label (as % of [1]) |
| [1] | 11 | 110 | 12 | 8 (66.7%) | 65 (59.1%) |
| [2] | 13 | 156 | 40 | 28 (70.0%) | 150 (96.2%) |
| [3] | 4 | 12 | 7 | 6 (85.7%) | 11 (91.7%) |
| [1, 2] | 15 | 210 | 52 | 36 (69.2%) | 198 (94.3%) |
| [1, 2, 3] | 13 | 156 | 40 | 28 (70.0%) | 148 (94.9%) |

It can be seen from Table 5 that the percentage of symmetrical links (i.e. Correctness) varied between 66.7% and 85.7%, which is a smaller variation range compared to Table 4 where the range is between 40% and 82.6%. This suggests that the direct use of ChatGPT can produce results that are more consistent across DSM models. Indeed, the engine components identified in [2] and [1, 2, 3] are identical and resulted in the same Correctness of 70.0% when ChatGPT was used directly. ChatGPT produced higher Correctness for [1] and [3] but lower Correctness for [2], [1, 2], and [1, 2, 3] when compared to the results in Table 4. Hence, it was inconclusive if ChatGPT can produce better Correctness based on the results of this work. Nevertheless, Completeness was found to be lower across all the settings examined when ChatGPT was used directly, suggesting that the use of Auto-DSM in populating DSM entries has an advantage.

The five settings of Auto-DSM and the direct use of ChatGPT were subsequently applied on the diesel engine components documented in [4] for evaluation. While Auto-DSM is primarily developed to support the analysis of organisation-specific design dependencies through the use of proprietary data, the objective here is to examine if the models created can be generalized and applied on other use cases such as [4]. Figure 6 shows the DSM reproduced from [4] with mechanical-static links coloured in red. For comparison, the six corresponding DSMs generated through Auto-DSM and ChatGPT based on the engine components from [4] are shown in Figure 7.

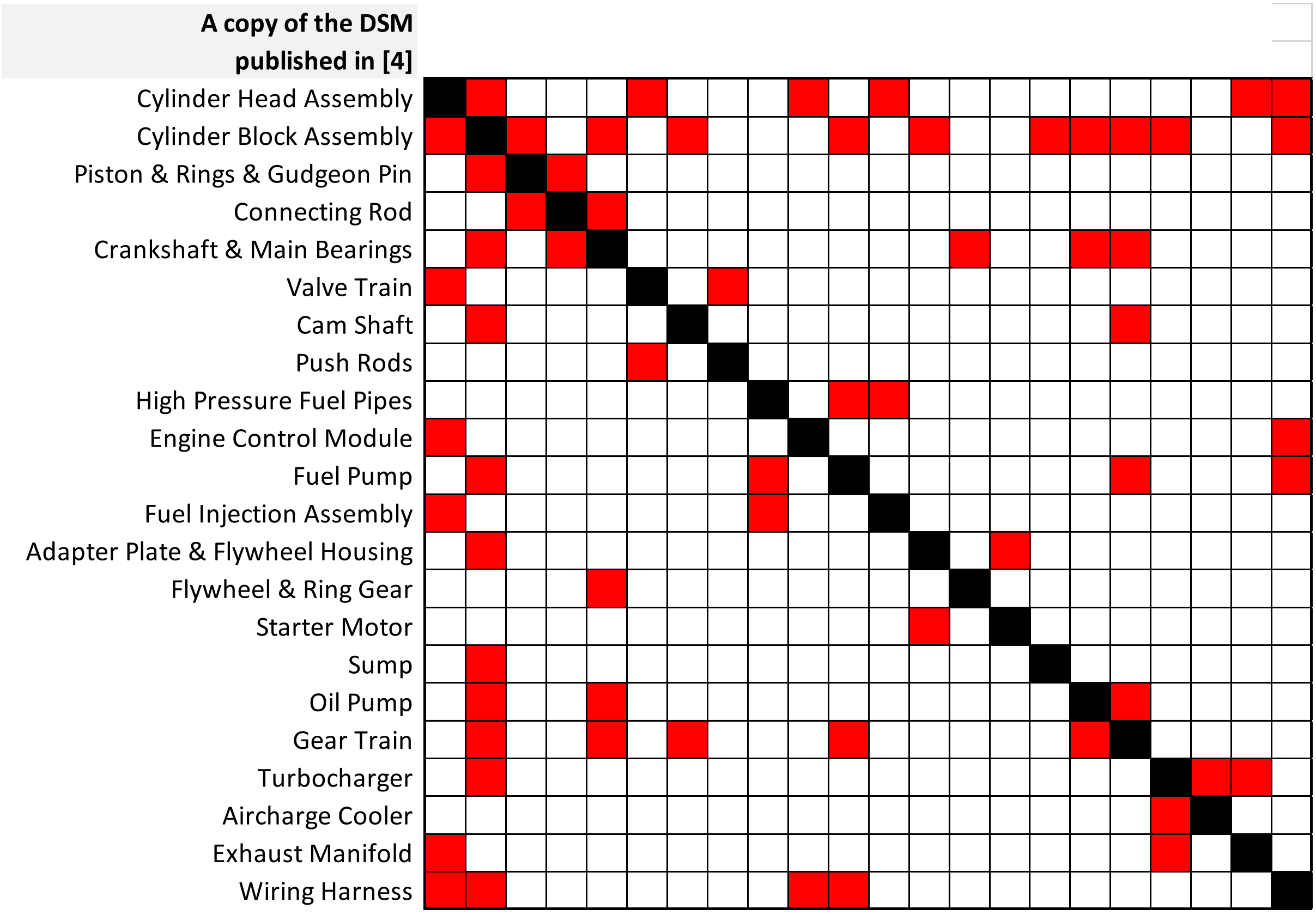

**Figure 6: A DSM of a diesel engine reproduced from [4] (see Keller et al. [2005]).**

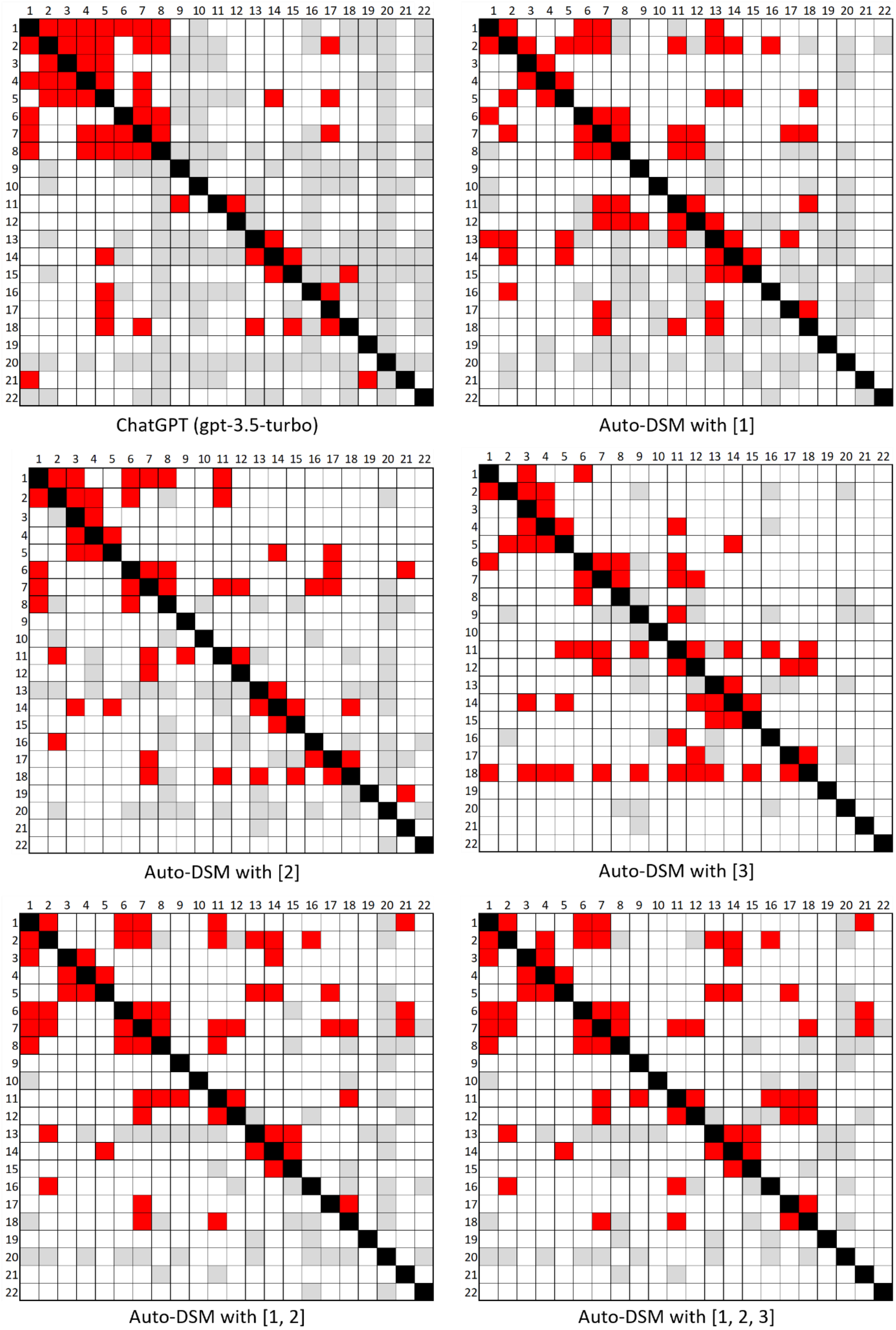

**Figure 7: DSMs generated with Auto-DSM and ChatGPT based on the engine components from [4], shown with component numbers from Table 3.**

Table 6 shows a comparison between the DSM entries reproduced from [4] and the entries generated through Auto-DSM and ChatGPT. As the DSM entries from [4] were manually elicited from experts, no value was computed for its percentage of symmetrical links found (i.e. Correctness) and percentage of entries with a useful label (i.e. Completeness). The highest Correctness value was produced by Auto-DSM with [1] at 82.5% while Auto-DSM with [2] produced the lowest at 59.3%. The direct use of ChatGPT on the components in [4] achieved a Correctness of 70.0%, which was inconclusive in determining whether ChatGPT can produce better Correctness compared to Auto-DSM. However, ChatGPT also produced a Completeness of 66.0%, which was lower across all the Auto-DSM settings. The result suggests that the use of Auto-DSM in populating DSM entries has an advantage over ChatGPT and is consistent with earlier findings from Table 4 and 5.

**Table 6: A comparison between the DSM entries generated from Auto-DSM and ChatGPT with [4].**

| | Number of | | | | | | |
|---|---|---|---|---|---|---|---|
| **Methods** | **Links found[4]** | **Symmetrical links found (as % of [4])** | | **Entries with a useful label[6] (as % of [5])** | | **Identical DSM entries as [4] (as % of [5])** | |
| DSM from [4] | 64 | 64 | (---) | 462 | (---) | 462 | (100.0%) |
| ChatGPT on [4] | 60 | 42 | (70.0%) | 305 | (66.0%) | 244 | (52.8%) |
| Auto-DSM with [1] on [4] | 63 | 52 | (82.5%) | 379 | (82.0%) | 315 | (68.2%) |
| Auto-DSM with [2] on [4] | 54 | 32 | (59.3%) | 398 | (86.1%) | 329 | (71.2%) |
| Auto-DSM with [3] on [4] | 57 | 36 | (63.2%) | 430 | (93.1%) | 357 | (77.3%) |
| Auto-DSM with [1, 2] on [4] | 59 | 44 | (74.6%) | 401 | (86.8%) | 335 | (72.5%) |
| Auto-DSM with [1, 2, 3] on [4] | 60 | 44 | (73.3%) | 409 | (88.5%) | 342 | (74.0%) |

[5]462 DSM entries in [4]

With reference to Table 6, Auto-DSM with [3] achieved the highest Completeness of 93.1% with 430 useful labelled entries. In addition, 357 of those entries are identical to the 462 DSM entries in [4]. This implies that 77.3% of the DSM entries in [4] could have been accurately generated using Auto-DSM. Table 6 also shows that combining input data in this work has a pooling effect as the exact number of identical DSM entries found when [1, 2] and [1, 2, 3] were used was within the highest and lowest number of identical DSM entries achieved by their constituent input data. This suggests that combining various sources of input data and having a bigger dataset can help avoid extreme results. Such an approach can be useful especially in cases where the quality of input data is unclear.

## 5. Discussions

The results from the diesel engine test case demonstrate that Auto-DSM can produce as high as 77.3% of entries identical to those generated by human experts. The results also show that Auto-DSM is sensitive to the input data used and combining different datasets can create a useful pooling effect, especially when the quality of input data is unclear. Although it was inconclusive if Auto-DSM can produce better Correctness compared to the direct use of ChatGPT, the results of this work show that Auto-DSM consistently produced better Completeness than ChatGPT.

As mentioned previously, the work of Dong and Whitney [2001] and the work of Wilschut et al. [2018] are two research articles that explicitly discuss automated DSM generation. Dong and Whitney [2001, pp 4] did not state their success metrics but indicated that five technical experts were consulted on the results and “Each expert reviewed the interactions in the DSM and agreed that most of the interactions captured were correct and reasonable. A few modifications were proposed. However, these proposed modifications were caused by the missing information during the construction of the DM” (DM refers to the Design Matrix in Axiomatic Design theory introduced by [Suh 1998]). The remarks indicate the need to create an accurate Design Matrix before the method can be used though the time needed to create one and the ability to automate the process were not reported. From this perspective, Auto-DSM has made a significant contribution towards automated DSM generation as it takes in documents directly without the need for pre-processing.

In the comparison with [Wilschut et al. 2018], Wilschut et al. state that their method identified all dependencies that were manually identified by experts in [Dijkstra 2016] but their results contained 14.7% more dependencies that were not previously identified. This implies that [Wilschut et al. 2018] can produce more correct entries than Auto-DSM. However, Wilschut et al. acknowledge that getting the data into the required structure can be time-consuming where the generation of a DSM with 45 elements took approximately 4 to 5 weeks and involved a review of roughly 800 pages of design documentation to prepare. As mentioned earlier, Auto-DSM took less than 4 minutes on a 11th Gen Intel(R) Core(TM) i7-1185G7 3.00GHz processor with 16.0 GB installed RAM to produce the DSM in Figure 4 (11 elements, exact time depends on internet connection). Hence, despite having a lower accuracy compared to [Wilschut et al. 2018], Auto-DSM has made a significant contribution in terms of the time required to generate a DSM.

Overall, the results from the diesel engine test case show promise and the no-code prototype of the workflow offers an opportunity for other researchers with no coding background to build upon. For instance, future research can explore the use of Auto-DSM on other product systems and document types. The no-code prototype can support organisations in generating a first draft DSM as well, especially when time and resources are tight and manual generation is not possible. Hence, this work has a practical impact as it provides a means to support the industry adoption of DSM.

## 6. Conclusions

This paper presents the Auto-DSM workflow, which uses a Large Language Model (LLM) to support the automation of DSM generation. The goal is to improve productivity as manual DSM generation can be time-consuming and costly. A prototype of the workflow was developed in this work and a test case involving a diesel engine was carried out to examine the feasibility of the proposed workflow. The results show promise with Auto-DSM reproducing 77.3% of DSM entries that are identical to those manually generated by experts. A no-code version of the prototype is available online to serve as a construct for future research and to support industry adoption. More work needs to be done to explore the use of Auto-DSM on other product systems and document types.